\documentclass[10pt,twocolumn,letterpaper]{article}

\usepackage{iccv}
\usepackage{times}
\usepackage{epsfig}
\usepackage{graphicx}
\usepackage{amsmath}
\usepackage{amssymb}
\usepackage{graphicx}
\usepackage{multirow}
\usepackage{caption}
\usepackage{booktabs}
\usepackage{subcaption}

\usepackage[pagebackref=true,breaklinks=true,letterpaper=true,colorlinks,bookmarks=false]{hyperref}

\iccvfinalcopy 

\def\iccvPaperID{xxxx} 

\ificcvfinal\fi

\newcommand{\dname}{D-LAYERS}
\newcommand{\mname}{LayersNet}
\def\Tabref#1{Table~\ref{#1}}

\usepackage{amsmath,amsfonts,bm}

\def\Figref#1{Figure~\ref{#1}}

\def\eqref#1{equation~\ref{#1}}
\def\Eqref#1{Equation~\ref{#1}}

\def\1{\bm{1}}

\def\va{{\bm{a}}}

\def\vf{{\bm{f}}}

\def\vn{{\bm{n}}}

\def\vq{{\bm{q}}}
\def\vr{{\bm{r}}}
\def\vs{{\bm{s}}}

\def\vv{{\bm{v}}}
\def\vw{{\bm{w}}}
\def\vx{{\bm{x}}}

\def\mE{{\bm{E}}}

\def\mV{{\bm{V}}}

\DeclareMathAlphabet{\mathsfit}{\encodingdefault}{\sfdefault}{m}{sl}
\SetMathAlphabet{\mathsfit}{bold}{\encodingdefault}{\sfdefault}{bx}{n}

\newcommand{\softmax}{\mathrm{softmax}}

\begin{document}

\title{Towards Multi-Layered 3D Garments Animation}

\author{Yidi Shao$^1$ \quad Chen Change Loy$^1$ \quad Bo Dai$^2$\\
$^1$S-Lab for Advanced Intelligence, Nanyang Technological University\\
$^2$Shanghai AI Laboratory
}

\twocolumn[{%
      \renewcommand\twocolumn[1][]{#1}%
      \vspace{-1cm}
      \maketitle
      \ificcvfinal\thispagestyle{empty}\fi
      \vspace{-1.6cm}
      \begin{center}
        \centering
        \includegraphics[width=0.97\textwidth]{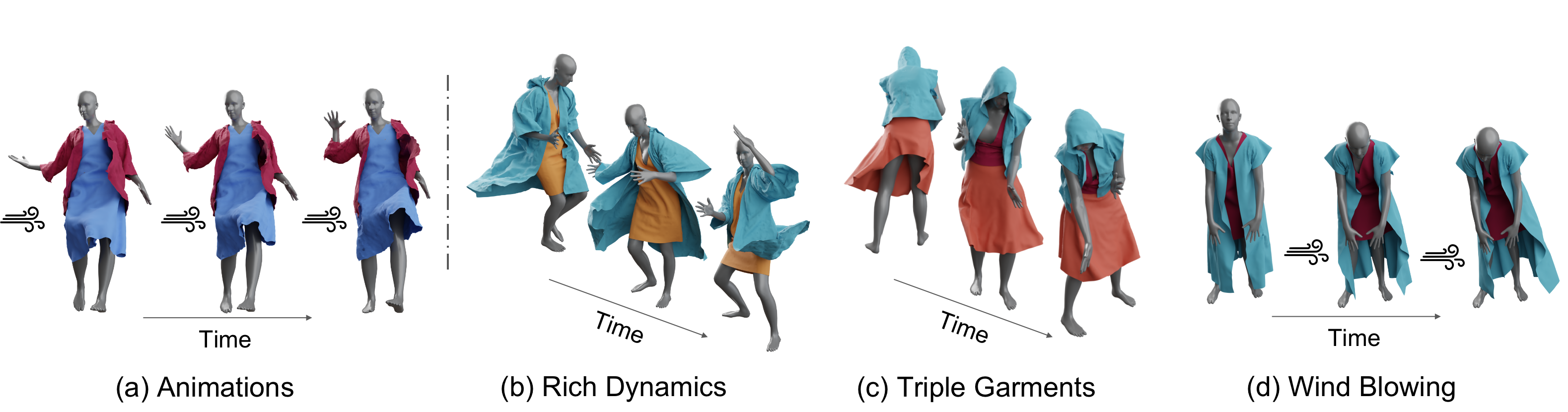}\vspace{0.1cm}
        \vskip -0.4cm
        \captionof{figure}{We propose \mname~with novel Rotation Equivarlent Transformation to animate garments in simulation manner.
        Our \mname~is able to animate multi-layered garments driven by various external forces, such as human bodies and wind as shown in (a). \mname~is powered by our proposed \dname, a novel large-scale 3D garment animation dataset involving realistic and challenging scenarios,
        as shown in (b)-(d).
        }
        \label{fig:teaser}
      \end{center}
    }]
\ificcvfinal\thispagestyle{empty}\fi

\begin{abstract}

Mimicking realistic dynamics in 3D garment animations is a challenging task due to the complex nature of multi-layered garments and the variety of outer forces involved. 
Existing approaches mostly focus on single-layered garments driven by only human bodies and struggle to handle general scenarios.
In this paper, 
we propose a novel data-driven method, called \textbf{\mname}, 
to model garment-level animations as particle-wise interactions in a micro physics system.
We improve simulation efficiency by representing garments as patch-level particles in a two-level structural hierarchy. 
Moreover, we introduce a novel \textbf{Rotation Equivalent Transformation} that leverages the rotation invariance and additivity of physics systems to better model outer forces. 
To verify the effectiveness of our approach and bridge the gap between experimental environments and real-world scenarios, 
we introduce a new challenging dataset, \textbf{\dname}, 
containing 700K frames of dynamics of 4,900 different combinations of multi-layered garments driven by both human bodies and randomly sampled wind. 
Our experiments show that \mname~achieves superior performance both quantitatively and qualitatively. 
We will make the dataset and code publicly available at
\url{https://mmlab-ntu.github.io/project/layersnet/index.html}
.

\end{abstract} 

\section{Introduction}

3D garment animation has been an active and important topic in computer graphics and machine learning, 
due to its great potential in various downstream tasks including virtual reality, virtual try-on, gaming and film production.
However, 
realistic 3D garment animation remains an open research problem 
due to the intrinsic challenge of modeling garments dynamics.

Specifically,
the dynamics of garments are jointly affected by both internal and external driving factors.
For internal factors,
while garments vary in topologies and materials, 
different topologies and materials result in drastically different dynamics.
Moreover,
in practice, 
humans usually wear multiple garments in a layered manner,
and such multi-layered garments further complicate the problem.
For example,
the rigid outer layer of a jacket can press against a softer inner dress, while the inner layer of a rigid t-shirt tries to maintain its shape against outer softer clothing.
As for external factors,
in addition to the movement of human body,
gravity, wind and friction also significantly influence the dynamics of garments in different ways.
Given the complexity of 3D garment animation,
previous approaches \cite{DBLP:journals/tog/WangSFM19, DBLP:journals/cgf/VidaurreSGC20, DBLP:conf/cvpr/MaYRPPTB20, DBLP:conf/cvpr/PatelLP20} tend to simplify the problem, 
considering only single-layered garments with the movement of human body being the only external driving factor.
Though being effective in such a simplified setting,
their applicability in real-life scenarios is significantly reduced.
Moreover,
they often resort to garment-specific designs,
which further limits their generality across garments with different topologies and materials.

In this paper,
we propose a novel data-driven method, \textbf{\mname}, for 3D garment animation,
which is inspired by the observation that 
although different driving factors, garment topologies and materials lead to significantly varying garments behaviors in the macro view,
at the micro level the dynamics of particles with same attributes share similarities.
Therefore, \mname~realizes a Transformer-based simulation system that utilizes the properties of rotation invariance and additivity to capture system dynamics via \textit{particle-wise interactions},
where garments, human body as well as other external factors are all represented by particles,
making \mname~agnostic to specific garment topology, the number of layered garments, and the set of considered external factors.
In practice,
we also adopt a two-level structural hierarchy in \mname,
where garments are made of patches, and patches consist of vertices of a fixed configuration.
Patches are thus treated as garments' basic particles,
and \mname~only needs to learn the interactions between patches,
resulting in a significant reduction of computational complexity.
To further improve the effectiveness of \mname,
we also propose a novel \textbf{Rotation Equivalent Transformation} to ease the modeling complexity of external factors.
Specifically,
while the external factors can influence garment particles in diverse directions,
the behaviors of interaction forces remain consistent in local canonical spaces,
which are under the directions of forces or the normals of obstacles' surfaces.
For instance,
the wind blows garments along the force directions,
while the meshes of human skin consistently push other objects outside of the body.
The proposed Rotation Equivalent Transformation thus transforms high-dimensional features to the local canonical space to reduce the redundant rotation information and capture interactions' semantics,
followed by transforming features back to the global space for aggregations.
In this way, it enables \mname~to effectively exchange semantics across multiple complicated external factors.

To verify the effectiveness of \mname~in more general cases and bridge the gap between experimental environments and real-world applications, 
we introduce a new challenging dataset called \textbf{\dname}, Dynamic muLti-lAYerEd gaRmentS, dataset. 
The dataset focuses on multi-layered garment animation driven by both the human body and wind. 
Multi-layered garments in \dname~are prepared as combinations of inner and outer clothes, each with different attribute values, such as bend stiffness and frictions. 
All garments on the same human body interact with each other, 
constrained by the laws of physics and simultaneously affected by the wind with randomly sampled direction and strength.
\dname~contains 4,900 different combinations of multi-layered garments and 700k frames in total,
with a maximum sequence length of 600 frames.
Experiments on \dname~demonstrate that \mname~outperforms existing methods and is more generalizable in complex settings.

Our contributions can be summarized as follows:
1) We propose a Transformer-based simulation method, \mname, 
with a novel rotation equivalent transformation for 3D garment animation that uses rotation invariance and additivity of physics systems to uniformly capture and process interactions among garment parts, different garments, as well as garments against driving factors. 2) We further propose \dname, a large-scale and new dynamic dataset for 3D garment animation. 
The dataset and code will be made publicly available.
\section{Related Work} \label{related_work}

\paragraph{Data-driven Cloth Model.}
Most existing approaches aim to estimate a function that outputs garment deformations for any input by learning a parametric garment model to deform corresponding mesh templates.
This is accomplished by modeling garments as functions of human pose \cite{DBLP:journals/tog/WangSFM19},
shape \cite{DBLP:journals/cgf/VidaurreSGC20},
pose-and-shape \cite{DBLP:conf/eccv/BerticheME20, DBLP:conf/iccv/BerticheMTE21, DBLP:conf/iccvw/TiwariB21},
motions \cite{DBLP:conf/cvpr/SantestebanTOC21},
garment type \cite{DBLP:conf/cvpr/MaYRPPTB20, DBLP:conf/cvpr/PatelLP20, DBLP:conf/cvpr/MaSYTB21},
and extended anchors beside human joints \cite{DBLP:conf/siggraph/PanMJTLS00M22}.
These approaches rely heavily on SMPL-based human models and and blend weights to animate garments according to registered templates, limiting generalization due to task-specific design.
To handle obstacles with arbitrary topologies,
N-Cloth \cite{DBLP:journals/cgf/LiTYHTYLM22} predicts garments deformations given the states of initial garments and target obstacles.
Other studies \cite{DBLP:conf/eccv/ShenLL20, DBLP:journals/corr/abs-2209-11449}
generate 3D garments based on UV maps.
SMPLicit \cite{DBLP:conf/cvpr/CoronaPAPM21} generates garments by controlling clothes' shapes and styles,
but intersection-free reconstruction is not guaranteed.

In contrast to existing methods, our \mname~animates garments by inferring garments' future positions through interactions between garment particles and other driving factors. Since driving factors are also represented by particles, garment animation simulates particle-wise interactions, which is shape-independent and generalizable to unseen scenarios.
A concurrent work \cite{DBLP:journals/corr/abs-2212-07242} adopts a Graph Neural Network (GNN)-based simulation network to model garment dynamics,
which uses interactions between adjacent vertices and distant vertices as edges,
resulting in redundant computational overhead.
On the contrary,
our proposed \mname~adopts a Transformer-based network
and model garment dynamics with patch-wise interactions,
so that the computational complexity is significantly reduced.
In addition,
the proposed novel rotation equivalent transformation further improves the effectiveness of \mname.

\begin{figure*}[t]
    \vspace{-4mm}
        \begin{center}
            \includegraphics[width=0.9\textwidth]{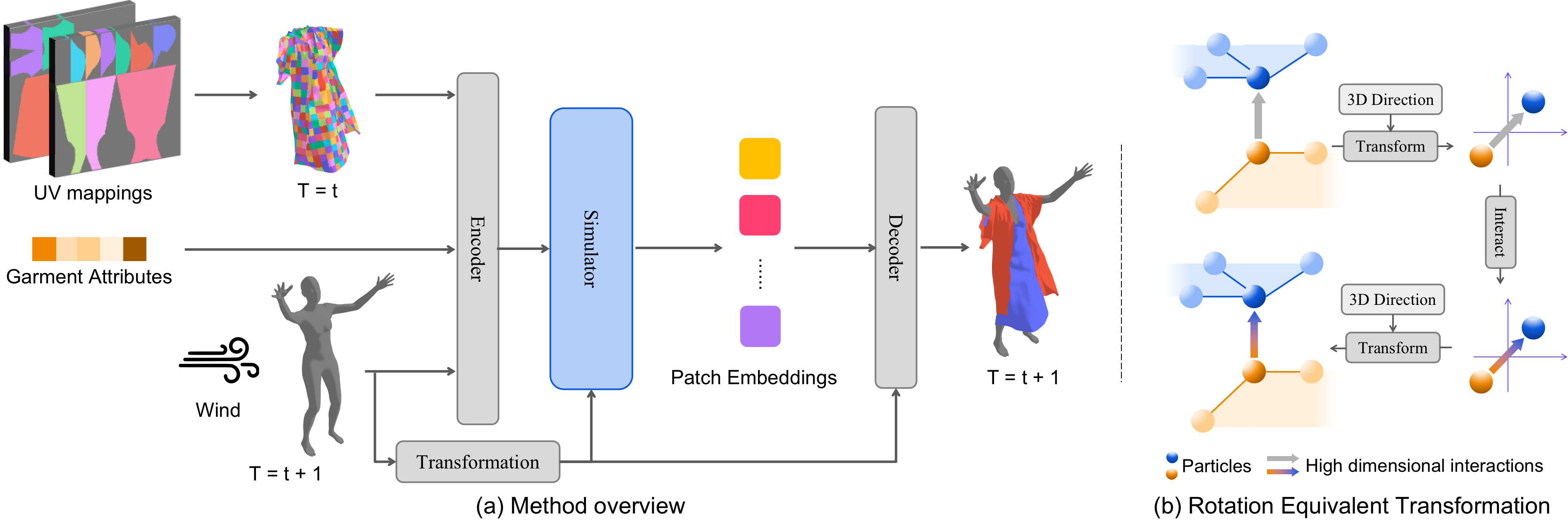}
        \end{center}
        \vspace{-4mm}
        \caption{\small{
            (a). Overview of \mname.
            Given driving factors at time $t+1$,
            i.e., the human body model and environmental wind in our study,
            \mname~animates target garments at time $t$
            and predicts the new states of garments at time $t+1$.
            While all objects are represented by particles.
            we establish a two-level structural hierarchy for garments,
            as shown on the top left of the figure,
            where garments are made of patches given the UV mappings.
            Then we encode the particles
            and model the interactions among them by a simulator,
            which outputs the embeddings for each patch.
            We apply a decoder to decode the vertices' dynamics at time $t+1$ given neighbor patches features.
            The Rotation Equivalent Transformation (RET) is applied to both simulator and decoder.
            (b) Key ideas about RET.
            In high dimensional space,
            we transform the interactions between garments' and external forces' particles into canonical spaces,
            which are defined by 3D directions, such as vertex normals, of corresponding external forces',
            and extract the semantics of interactions,
            followed by the transformation back to the shared hidden space for aggregation.
            The high dimensional transformation is calculated by our rotation network,
            which converts rotation matrix in 3D space to hidden space.
        }}
        \vspace{-2mm}
        \label{fig:overview}
\end{figure*}
\noindent\textbf{Rotation Invariant Neural Network.}
Many existing approaches \cite{DBLP:journals/corr/abs-1802-08219, DBLP:conf/nips/AndersonHK19, DBLP:conf/nips/FuchsW0W20, DBLP:conf/nips/GasteigerBG21, DBLP:conf/nips/GasteigerYG21, DBLP:conf/iclr/BrandstetterHPB22, DBLP:conf/iclr/0059WLLZOJ22} adopt spherical harmonics to encode higher-order interactions and achieve SE(3)-equivariance. These approaches focus on extracting and propagating rotation-invariant features through different layers. In contrast, while \mname~is motivated by the rotation equivalent property of physics systems, we aim to rotate high-dimensional features into local canonical space using the mapped rotation matrix from 3D space to eliminate rotation effects and model interactions involving outer forces. We then rotate the learned features back to the shared hidden space for aggregation.

\noindent\textbf{3D Garment Datasets.}
Existing 3D garment datasets are generated either synthetically \cite{DBLP:conf/iccv/PumarolaSCSM19, DBLP:conf/cvpr/PatelLP20, DBLP:conf/cvpr/SantestebanTOC21, DBLP:conf/eccv/BerticheME20} or from real-world scans \cite{DBLP:conf/cvpr/ZhangPBP17, DBLP:conf/iccv/ZhengYWDL19, DBLP:conf/cvpr/MaYRPPTB20, DBLP:conf/eccv/TiwariBTP20, DBLP:journals/corr/abs-2204-13686}. 
Synthetic datasets such as 3DPeople!\cite{DBLP:conf/iccv/PumarolaSCSM19}, TailorNet~\cite{DBLP:conf/cvpr/PatelLP20}, and Cloth3D~\cite{DBLP:conf/eccv/BerticheME20}, mostly contain single-layered 3D garment models, and while some datasets have multiple garments, there are very few overlapping areas among different cloth pieces~\cite{DBLP:conf/iccv/BhatnagarTTP19}. Layered-Garment Net~\cite{ACCV2022layered} proposes a static multi-layered garments dataset in seven static poses for 142 bodies to generate layers of outfits from a single image, but the garments mostly consist of skinning clothes that do not follow physics laws, and interpenetration is solved by simply forcing penetrated vertices out of inner garments.
To our knowledge, \dname~is the first dataset to include dynamic multi-layered 3D garments. The different layers of garments have distinct attributes and interact with each other, following the laws of physics. Furthermore, we introduce wind as an extra driving factor to animate the garments, adding complexity to their dynamics given similar human movements. Our dataset provides all the necessary 3D information, allowing for easy generalization to other tasks, such as reconstructions from a single image.

\noindent\textbf{Physics Simulation by Neural Network.}
Learning-based methods for physics simulation can be applied
to different kinds of representations, e.g., approaches for grid representation \cite{thuerey2020deep,wang2020towards},
meshes \cite{nash2020polygen,qiao2020scalable,weng2021graphbased,pfaff2021learning},
and particles \cite{li2019learning,Ummenhofer2020Lagrangian,pmlr-v119-sanchez-gonzalez20a, DBLP:conf/eccv/ShaoLD22}.
Some methods adopt GNN \cite{li2019learning, pmlr-v119-sanchez-gonzalez20a, pfaff2021learning}.
Another approach \cite{DBLP:conf/nips/LiangLK19} focuses on accelerating gradient computation for collision response, serving as a plug-in for neural networks.
A recent method, TIE~\cite{DBLP:conf/eccv/ShaoLD22}, applies Transformer with modified attention to recover the semantics of interactions. 
Our \mname~is inspired by TIE in the notion of modeling particle-wise interactions, thus inheriting the appealing properties of being topology-independent and easy to generalize to unseen scenarios.
Different from TIE, we establish a two-level hierarchy structure for garments,
which are made of deformable patches.
We further propose a rotation equivalent transformation to extract canonical semantics under different local coordinates in high dimensions to cope with complex outer forces.

\section{Methodology} \label{method}

\Figref{fig:overview} presents an overview of our proposed method, \mname, 
which aims to animate garments faithfully, regardless of their topology and driving factors.
In our case, these factors include rigid human bodies and winds. 
To achieve this, we introduce a patch-based garment model, 
which enables us to simulate the garment animation in a particle-wise manner. 
The main novelty of \mname~lies in our use of the properties of rotation invariance and additivity of physics systems. 
Specifically, 
we propose a Rotation Equivalent Transformation 
that employs a rotation invariant attention mechanism and a rotation mapping network 
to enable the communication and aggregation of semantics from different canonical spaces in a unified manner. 
In the following sections, 
we describe our particle simulation formulation for garment animation, the patch-based garment model, and the Rotation Equivalent Transformation.

\subsection{\mname}
\paragraph{Problem Formulation.}

We denote each mesh at time $t$ by $M^t=\{\mV^t, \mE^M, \mE^W\}$,
where $\mV^t = \{\vx^t_i, \dot{\vx}^t_i, \ddot{\vx}^t_i\}_{i=1}^N$
are the vertices' positions, velocities, and accelerations,
and $\mE^M$ denote the mesh edges.
$\mE^W$ are the world space edges \cite{pfaff2021learning},
where we dynamically connect node $i$ and node $j$
if $\vert \vx^t_i -  \vx^t_j \vert < R$,
excluding node pairs already exist in the mesh.
In a particle-based system,
each mesh is represented by particles,
which correspond to the vertices of the mesh.
During simulation,
particle $i$ and particle $j$ will interact with each other
only if an edge $e_{ij}$ connects them, where $e_{ij}\in \mE^M \cup \mE^W$.
The interactions guided by $\mE^M$ enable learning internal dynamics of mesh,
while interactions indicated by $\mE^W$ serve to compute external dynamics
such as collisions.

We adopt abstract particles to represent the garments' attributes and the wind.
Specifically, we use $\va_g$ to denote each garment's attribute,
such as friction and stiffness, and $\vw^t$ to denote the wind.
Since the wind has constant strength in the whole 3D space,
we use the quaternion rotation $\bm{\eta}^t$
and the strength $s^t$ to represent the wind as
$\vw^t = \{\bm{\eta}^t, s^t\}$.
In this way,
given the human body and wind at $t+1$
as well as their previous $h$ states, 
we predict the garments' states at time $t+1$
given the current states at $t$ and corresponding previous meshes $\{M^{t-1}, \cdots, M^{t-h}\}$.
In practice, we choose $h=1$ in all experiments.
Our approach can be described as:
\begin{eqnarray}
    \hat{\mV}^{t+1}_{g} &   =   &   \Gamma(\va_g, \{M^{t-i}_g, M^{t+1-i}_b, \vw^{t+1-i}\}^h_{i=0}),
\end{eqnarray}
where $M^t_g$ and $M^{t+1}_b$ are the meshes of garments and human body, respectively,
$\Gamma(\cdot)$ is the simulator and runs recursively during predictions,
and $\hat{\mV}^{t+1}_{g}$ is the garment's new vertices' states
at time $t+1$.
We adopt an encoder to embed the inputs into hidden space, and an decoder to decode the hidden features back to the 3D states.

\noindent\textbf{Patch-based Garment Model.}
Since garments are composed of hundreds and thousands of particles,
modeling interactions between densely connected particles inevitably leads to significant computational overhead.
To reduce the number of interactions,
we establish a two-level structural hierarchy for garments
and represent each garment by patches,
which consist of vertices of a fixed configuration.
Patches are treated as special particles and interact with each other during simulations instead of densely connected vertices.
Patch modeling holds several advantages.
First, as basic units to represent garments, patches are topology independent.
By modeling the dynamics of each patch, our model is more flexible and generalizable to unseen garments.
Second, instead of simulating each vertex in a mesh, simulating patches significantly reduces the computational overhead, especially when the mesh is of high fidelity.

Formally, we find a mapping $\rho(\cdot)$ to map the vertex-based mesh to patch-based representation by:
\begin{eqnarray}
    P^t_g   &   =   &   \rho(M^t_g),
\end{eqnarray}
where $P^t_g=\{\mV^t_p, \mE^M_p, \mE^W_p\}$.
The patches' states $\mV^t_p$ are the averaged vertices' states within the patches,
and $\mE^W_p$ are computed given $\mV^t_p$.
The mapping $\rho(\cdot)$ is based on the garments' UV maps as shown in \Figref{fig:overview}.
In this way,
our method can be updated as:
\begin{eqnarray}
    \hat{\mV}^{t+1}_{g} &   =   &   \Gamma(\va_g, \{P^{t-i}_g, M^{t+1-i}_b, \vw^{t+1-i}\}^h_{i=0}).
\end{eqnarray}

\noindent\textbf{Rotation Equivalent Transformation.}
Physics systems used for garment simulations possess two essential properties: rotation invariance and additivity. The rotation equivariance property states that the interactions' effects between objects remain the same regardless of the objects' rotations, while the additivity property implies that the total influence towards a particle equals the summation of each component's influence. By exploiting these two properties, we can segregate the impact of directed forces, such as forces brought by complex surface human bodies and directed wind, into individual interactions, solve them within their canonical space, and then aggregate the results. We assume the z-axis of the canonical space is the direction of human model vertex normal or the wind field, while the remaining two axes can be randomly selected, thanks to rotation invariance. To ensure that our Transformer-based model equivalently pays attention to features under different rotations,
we apply decentralization and normalization for attention keys (\eqref{eq:edge}),
and propose a rotation-invariant attention mechanism
\begin{eqnarray}
    \vq_i&=&W_q \vv_i, \qquad \vr_i = W_r \vv_i,  \qquad \vs_i=W_s \vv_i,\\
    \vf_{i,j} &   =   &   \frac{\vr_i+\vs_j-\bm{\mu}_{\vr_i, \vs_j}}{\sigma_{\vr_i, \vs_j}}, \label{eq:edge}\\
    \omega_{ij} &   =   &  \softmax(\vq_i^\top \vf_{i,j}), \label{eq:attn}
\end{eqnarray}
where $\vv_i$ is state token, $\vq_i$ is query token, $\vr_i$ is receiver token and $\vs_j$ is sender token,
$W_q, W_r, W_s$ are trainable parameters.
$\bm{\mu}_{\vr_i, \vs_j} = (\vr_i+\vs_j)/2$ is the mean vector of $\vr_i$ and $\vs_j$,
while $\sigma_{\vr_i, \vs_j}$ is the corresponding standard deviation.
The choices of $\bm{\mu}_{\vr_i, \vs_j}$ and $\sigma_{\vr_i, \vs_j}$ ensure \Eqref{eq:edge} is rotation equivariant,
which decentralizes the feature vectors and normalizes them by the averaged L2 distance towards the center.
The proof can be found in supplementary materials.
\Eqref{eq:edge} can be further simplified as:
\begin{eqnarray}
    \vf_{i,j} &   =   &   \frac{\vr_i+\vs_j}{\Vert \vr_i - \vs_j\Vert}. \label{eq:edge_simp}
\end{eqnarray}

To directly extract rich semantics from high-dimensional spaces for interactions rather than 3D space, and rotate them into potential canonical space,
we propose a rotation network to model high-dimensional rotations given the corresponding 3D rotations.
Specifically,
for each human body vertex $\vv_{b_j}$,
we calculate the rotation matrix $R_{b_j} \in \mathbb{R}^{3\times3}$ that transforms the 3D world space coordinates into local coordinates,
where the z-axis is the normal $\vn_{b_j}$ of $\vv_{b_j}$.
Since the physics system is rotation invariant,
we can randomly sample a unit vector orthogonal to $\vn_{b_j}$ as x-axis, and get the y-axis unit vector through cross product.
To find the corresponding rotation matrix in the $l$-th layer with $d$ dimension,
we design a rotation network $\phi^l(\cdot):\mathbb{R}^{3\times 3} \to \mathbb{R}^{d\times d}$ as:
\begin{eqnarray}
    \phi^l(R)   &=& W^l_R R (W^l_R)^\top,
\end{eqnarray}
\begin{equation}
    \text{s.t.} \quad W^l_R (W^l_R)^\top = I, \quad(W^l_R)^\top W^l_R = I, \label{eq:rot_cond}
\end{equation}
where $W^l_R \in \mathbb{R}^{d\times 3}$ is the learnable parameter.
\Eqref{eq:rot_cond} ensures the rotation matrix in hidden space satisfying the property $\phi^l(R)(\phi^l(R))^\top=I$.
The interactions between $i$-th garment patch and its neighbor human body vertex $b_j \in \mathcal{N}^b_{i}$ as well as the rest patches $k \in \mathcal{N}^p_{i}$ at $l$-th layer can be written as:
\begin{eqnarray}
    \vf^R_{i,b_j} &   =   &   \psi(\phi(R_{b_j})\vf_{i, b_j}), \\
    \vv^\prime_{i} &   =   &   \sum_{b_j}\omega_{ib_j} (\phi(R_{b_j}))^\top \vf^R_{i,b_j} 
    + \sum_{k}\omega_{ik} \vf_{i,k},\label{eq:rot_agg} 
\end{eqnarray}
where $\vv^\prime_{i}$ is the updated state token for $i$-th patch, and $\psi(\cdot)$ is multi-layer perception in practice. The first term in \Eqref{eq:rot_agg} rotates the interaction features $\vf^R_{i,b_j}$ from different canonical space back to the shared hidden space before aggregation.
For gravity and wind, the directions of the forces are used to calculate the rotation matrices.

Finally,
to recover the details of $k$-th vertex,
we utilize its neighbor patches $p_i\in\mathcal{N}^p_{k}$
and the nearest point on human body indexed by $b_j$ for decoding as follows:
\begin{eqnarray}
    \vv^{R}_{p_i}  &=& \phi(R_{b_j})\vv_{p_i}, \qquad \vv^{R}_{b_j}=\phi(R_{b_j})\vv_{b_j},\\
    {\bm{\alpha}}^{R, t+1}_{k} &   =   &   \frac{1}{N_k}\sum_{p_i}g([R_{b_j}(\bar{\vx}^{t}_k-\bar{\vx}^{t}_{p_i}), \vv^{R}_{p_i}, \vv^{R}_{b_j}]), \label{eqn:decode}\\
    {\bm{\beta}}^{t+1}_{k} &   =   &   \Delta t\cdot(R_{b_j})^\top {\bm{\alpha}}^{R, t+1}_{k}+\bar{\bm{\beta}}^{t}_k, \\
    {\vx}^{t+1}_{k} & = &     \Delta t\cdot{\bm{\beta}}^{t+1}_{k}+\bar{\vx}^{t}_k,
\end{eqnarray}
where we first rotate the patches' features $\vv_{p_i}$, human body vertex features $\vv_{b_j}$, and the ground truth relative positions $\bar{\vx}^{t}_k-\bar{\vx}^{t}_{p_i}$ at time $t$ before concatenation. We average the output of decoder $g(\cdot)$ as the predicted 3D acceleration ${\bm{\alpha}}^{R, t+1}_{k}$, which is further transformed back to global 3D coordinates to compute the velocity ${\bm{\beta}}^{t+1}_{k}$ and position ${\vx}^{t+1}_{k}$ at time $t+1$. 
$\Delta t$ is the time interval between each frame.

\subsection{Training Details} \label{sec:train}
To train our simulation-based model,
we first apply a standard mean square error (MSE) loss as:
\begin{eqnarray}
    \mathcal{L}^{t+1}_{m, *} &   =   &   \frac{1}{N}\sum_i \Vert {\vx}^{t+1}_i-\bar{\vx}^{t+1}_i \Vert_2^2,
\end{eqnarray}
where $\{\vx_i^{t+1}\}_{i=1}^{N}, \{\bar{\vx}_i^{t+1}\}_{i=1}^{N}$ are the predictions and ground truths at time $t+1$ respectively.
We penalize the MSE loss on both garment vertices' positions $\mathcal{L}^{t+1}_{m, g}$ and the center of patches' positions $\mathcal{L}^{t+1}_{m, p}$ together as $\mathcal{L}^{t+1}_{m}=\mathcal{L}^{t+1}_{m, g}+\mathcal{L}^{t+1}_{m, p}$.

We adopt a loss term for the garment vertex normal to maintain the smoothness
and consistency as:
\begin{eqnarray}
    \mathcal{L}^{t+1}_{n} &   =   &   \frac{1}{N_v}\sum_i \Vert {\vn}^{t+1}_i-\bar{\vn}^{t+1}_i \Vert_2^2,
\end{eqnarray}
where ${\vn}^{t+1}_i$ and $\bar{\vn}^{t+1}_i$ are the vertex normals
for prediction and ground truth, respectively.

To further reduce the collision rates between garments and human bodies,
as well as between different layers of garments,
we adopt collision loss:
\begin{eqnarray}
    \mathcal{L}^{t+1}_{c} =  \frac{1}{N_{c}}\sum_i \max\left(d_{\epsilon}-({\vx}^{t+1}_i-\vx^{t+1}_a)  \vn^{t+1}_a, 0 \right)^{2}, \label{eqn:closs}
\end{eqnarray}
where $\vx^{t+1}_a$ is the nearest anchor point to ${\vx}^{t+1}_i$,
$N_c$ is the number of collided vertices,
and $d_{\epsilon}$ is the minimum distance of penetration.
The collision loss between garments and ground truth human bodies, as well as between predictions of layers of garments,
can be denoted as $\mathcal{L}^{t+1}_{c, \bar{b}}$ and $\mathcal{L}^{t+1}_{c, g}$ respectively.
Thus,
for predictions at time $t+1$,
our training loss is written as:
\begin{eqnarray}
    \mathcal{L}^{t+1} =  \lambda_{m} \mathcal{L}^{t+1}_{m} + \lambda_{n} \mathcal{L}^{t+1}_{n} + \lambda_{b} \mathcal{L}^{t+1}_{c, \bar{b}}+\lambda_{g} \mathcal{L}^{t+1}_{c,g}.
\end{eqnarray}

During training,
we randomly rollout $T_{n}$ steps without gradient given inputs at time $t-T_{n}$,
which aims to add noise from the model itself. We only back-propagate gradients from one-step predictions on time $t+1$.

\section{\dname~Dataset} \label{dataset}
\begin{table*}[ht]
\vspace{-6mm}
\setlength{\tabcolsep}{3pt}
\caption{
\small{
We display the influence of multi-layered garments and wind with different combinations for garment animations.
We list the components of different splits and models' corresponding Euclidean errors (mm) below.
Notice that in our \dname,
all objects are scaled up 10 times than real-world size.
We sample four splits from our dataset:
inner garments are tight clothes without wind (\textbf{T});
inner garments are tight clothes with strong wind (\textbf{T+W});
inner garments are loose clothes without wind (\textbf{L});
inner garments are loose clothes with strong wind (\textbf{L+W}).
The models marked by $*$ are trained and tested on the inner garment only.
Notice that MGNet has worse generalization abilities due to garment-specific design. 
\mname~achieves superior and robust performance in most cases especially those with multi-layered garments.}}
\vspace{-4mm}
\label{tbl:splits}
\begin{center} \small 
\begin{tabular}{lcclccccc}
\toprule
\multirow{2}{*}{\bf Splits}	&\multicolumn{3}{c}{\bf{Components}}    & \multicolumn{3}{c}{\bf Methods on Inner Garment} & \multicolumn{2}{c}{\bf Methods on Layered Garments}  \\ \cmidrule(lr{.75em}){2-4}\cmidrule(lr{.75em}){5-7}\cmidrule(lr{.75em}){8-9}
						&InnerCloth		&OuterCloth 		&WindStrength		&DeePSD$^*$\cite{DBLP:conf/iccv/BerticheMTE21}         &MGNet$^*$\cite{DBLP:journals/corr/abs-2209-11449}  &\mname$^*$(Ours)   &DeePSD\cite{DBLP:conf/iccv/BerticheMTE21}          &\mname(Ours)
\\ \cmidrule(lr{.75em}){1-4}\cmidrule(lr{.75em}){5-7}\cmidrule(lr{.75em}){8-9}
\bf T		    & Jumpsuit		& Jacket	        & $\le50$   &225.3$\pm$106.4    &5219.2$\pm$1565.8    &\textbf{220.0$\pm$195.3}		&1072.6$\pm$694.7   & \textbf{489.9$\pm$447.7}\\
\bf T+W 		& Jumpsuit		& Jacket	        & $>250$	&\textbf{239.5$\pm$103.9} &5186.8$\pm$1754.8  &258.0$\pm$312.8		        &1121.0$\pm$731.7	& \textbf{508.5$\pm$562.7}\\
\bf L           & Dress		    & Jacket	        & $\le50$	&501.3$\pm$300.1		&4432.7$\pm$1438.0		&\textbf{344.7$\pm$244.5}		&887.8$\pm$460.5	& \textbf{378.0$\pm$293.0}\\
\bf L+W         & Dress	        & Jacket	        & $>250$   &577.5$\pm$373.9    &4595.0$\pm$1215.2    &\textbf{347.4$\pm$288.9}       &1083.1$\pm$492.2   & \textbf{467.8$\pm$403.7}\\\bottomrule
\end{tabular}
\end{center}
\vspace{-4mm}
\end{table*}

Most existing datasets are limited to single-layered garments driven solely by human bodies. Different garments, such as the upper T-shirt and lower pants, rarely interact with each other. Consequently, the problem can be easily solved by 
modeling garments as functions of human bodies with single-layered outfits predictions \cite{DBLP:conf/cvpr/PatelLP20, DBLP:conf/iccv/BerticheMTE21}.
Collecting a real-world dataset with dynamic multi-layered garments and outer forces is expensive and usually contains noisy artifacts, such as interpenetration \cite{DBLP:conf/cvpr/MaYRPPTB20} between scanned clothes and estimated SMPL-based human bodies 
, while synthetic data are easier to obtain and can provide more accurate dynamics in most cases, particularly for multi-layered clothes with narrow gaps.
With this motivation, we generated \dname~using a simulation engine and Blender\footnote{\url{https://www.blender.org/}}, making it the first dynamic multi-layered garments dataset that considers the wind factor in addition to human bodies.

We construct our dataset by first collecting garment templates from SewPattern~\cite{DBLP:conf/nips/KorostelevaL21}, which includes various types of garments such as jackets with hoods and dresses with waist belts. We then generate multi-layered combinations of outer and inner-layer clothes. Each multi-layered garment combination is draped onto an SMPL human body model \cite{DBLP:journals/tog/LoperM0PB15}, followed by a warm-up simulation in Blender to resolve interpenetrations. Finally, we simulate the dynamics of the garments given human motion sequences from CMU MoCap in AMASS \cite{DBLP:conf/iccv/MahmoodGTPB19} and sampled winds. To preserve high-frequency details in Blender, we scale up the human and garment meshes ten times their real-world size before simulation.
Given the availability of the 3D meshes and attributes of garments, as well as the detailed scene settings for each sequence, \dname~offers the potential to extend to other formats of data and support explorations of alternative topics such as optical flow estimations, 3D reconstructions from images, and physics parameter estimations. Supplementary materials provide additional details on the key settings in \dname. Here, we highlight the two main settings:

\noindent\textbf{Multi-layered Garments.}
Each multi-layered outfit in \dname~consists of an inner and outer outfit with different garment attributes, such as mass, stiffness, and friction, leading to more diverse and flexible dynamics. For example, the outer garments can be softer or more rigid than the inner outfit. The outer outfit in our dataset is either a jacket or a jacket with a hood, providing a clear view of interactions from the inside and outside. Inner outfits refer to whole-body outfits, such as dresses, jumpsuits, and t-shirts with pants or skirts. We generate 4,900 combinations of multi-layered garments, which includes 9,872 different garments in total. The garment templates are of high fidelity, with vertices ranging from 5,000 to more than 15,000 for each garment, enabling us to capture more details in simulation.

\noindent\textbf{Wind.}
Most existing datasets simplify real-life scenarios by driving garment animation solely through human bodies. To enrich the simulation settings and enable researchers to explore garment animation driven by multiple factors, we introduce randomly sampled wind in \dname. Wind is a common and prominent force field that influences garment dynamics in the real world. To simulate wind in our dataset, we randomly select several intervals of frames in each sequence and apply winds with varying directions and strengths as force fields. The directions and strengths are uniformly sampled from 0 to 400 in Blender. Within each interval, we assume that the wind affects the entire 3D space, with the direction and strength remaining constant.

\section{Experiments} \label{experiment}
\begin{table*}[t]
    \vspace{-10mm}
    \setlength{\tabcolsep}{4pt}
    \caption{
    \small{
    Euclidean error (mm) on sampled \dname~with maximum sequence length of 35 frames.
    The collision rates between different layers of garments are shown under \textbf{L-Collision},
    while the collision rates between garments and human bodies are shown under \textbf{H-Collision}.
    Models trained with collision loss $\mathcal{L}_{c,b}$, $\mathcal{L}_{c, g}$ are marked by +.
    Our \mname~achieves superior results in all cases.
    }}
    \vspace{-4mm}
    \label{tbl:full}
    \begin{center} \small
    \begin{tabular}{lccccc}
    \toprule
    \bf Methods	    & \bf Jacket & \bf Jacket + Hood  & \bf Dress   &\bf Jumpsuit    &\bf Skirt
    \\ \midrule
    DeePSD          & 1385.3$\pm$886.6	& 1087.8$\pm$564.5	& 736.8$\pm$466.6 	& 535.2$\pm$224.7 & 1107.3$\pm$769.2 \\
    DeePSD+         	& 1830.1$\pm$803.3	& 1566.0$\pm$527.1	& 1333.0$\pm$349.2 	& 1219.0$\pm$186.8	& 1194.7$\pm$311.2\\
    \mname(Ours)	& 571.9$\pm$451.9	& 493.9$\pm$354.2 & 397.2$\pm$342.2 	& 264.0$\pm$200.2    &301.3$\pm$79.3\\
    \mname+(Ours)	& \textbf{567.3$\pm$425.5}	& \textbf{491.4$\pm$361.3} & \textbf{379.1$\pm$299.7} 	& \textbf{260.1$\pm$222.2}    &\textbf{299.5$\pm$92.3}\\
    \midrule
    \bf{Methods}	& \bf Pants  & \bf T-shirt & \bf Overall & \bf L-Collision(\%)    & \bf H-Collision(\%)\\
    \cmidrule(lr{.75em}){1-4}\cmidrule(lr{.75em}){5-6}
    DeePSD          & 498.8$\pm$109.5	& 613.1$\pm$338.2	& 1049.8$\pm$549.7 	& 10.11$\pm$5.31 & 23.89$\pm$7.89 \\
    DeePSD+			& 1185.7$\pm$213.3	&1202.9$\pm$233.6	& 1563.4$\pm$486.8	& 8.78$\pm$5.12	 	&19.47$\pm$6.38\\ 
    \mname(Ours)	& 234.4$\pm$206.3	& 273.3$\pm$169.0	& 472.8$\pm$343.5   & \textbf{3.13$\pm$2.22}	& 10.68$\pm$4.53   \\
    \mname+(Ours)	& \textbf{200.9$\pm$140.1}	& \textbf{267.8$\pm$189.6}	& \textbf{467.2$\pm$330.7}	& 3.77$\pm$2.60  	&\textbf{2.16$\pm$1.46} \\
    \bottomrule
    \end{tabular}
    \end{center}
    \vspace{-2mm}
\end{table*}
\begin{figure*}[!h]
    \vspace{-4mm}
        \begin{center}
            \includegraphics[width=0.9\textwidth]{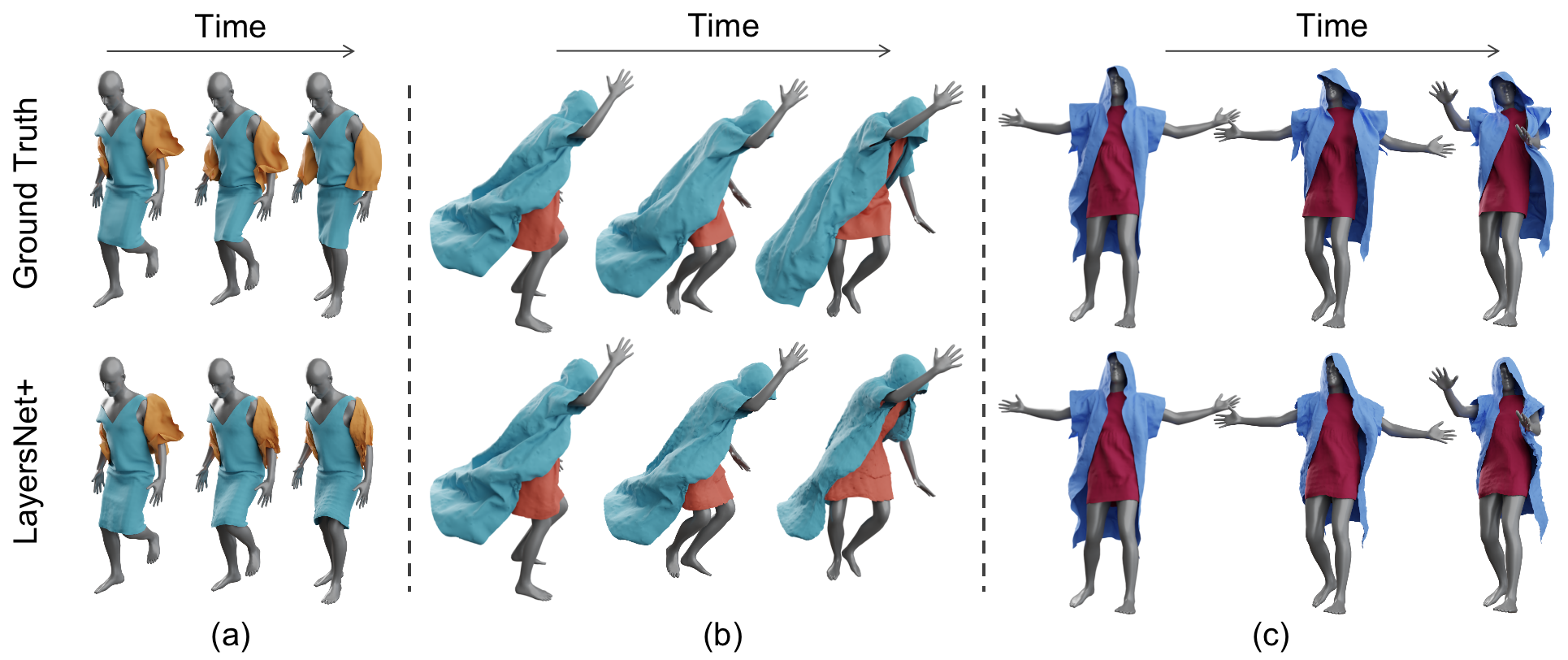}
        \end{center}
        \vspace{-6mm}
        \caption{\small{
        Qualitative results by our \mname+.
        Sequences in test set are mutually exclusive from training set samples.
        \mname+ is capable of generalizing to unseen scenarios with faithful and realistic rollouts
        in terms of vivid dynamics constrained by physics laws,
        low rates of garment-to-body and garment-to-garment interpenetration.
        (a). \mname+ animates a human walking downstairs with garments falling off the shoulders.
        The predictions of \mname+ shows the inertia of the jacket.
        (b). A human model climbs up and sits down.
        \mname+ rollouts the complex dynamics between dress and outer jacket,
        which is pushed aside by inner dress driven by human body.
        (c). A human model walks towards his left and hugs.
        While \mname+ is able to describe the inertia of the jacket which tends to move towards right,
        the jacket is hindered by the left side of dress and stopped by the left arm.
        }
        }
        \vspace{-4mm}
        \label{fig:qualitative}
\end{figure*}
\subsection{Baselines}

We implement DeePSD \cite{DBLP:conf/iccv/BerticheMTE21}
and MGNet \cite{DBLP:journals/corr/abs-2209-11449} as our baselines.
MGNet is a standard garment-specific model,
while DeePSD models garments as functions of human bodies
and achieves state-of-the-art performance in terms of 3D garment animations.
DeePSD claims to support the animations of multi-layered garments.
We make the following extensions to baselines:
1. We add wind as extra inputs;
2. We add the collision loss
between different layers of garments. The second extension only applies to multi-layered clothes setting of DeePSD, and does not apply to MGNet due to its specific design for single-layered clothes.
All models are trained with ten epochs.
We do not apply any post-processing for both the training and prediction stages.
During the evaluation, we calculate the mean of Euclidean errors for each frame, then average the errors across all the frames within each sequence.
The final results are the mean of errors from all sequences.

\subsection{Garment Animations on \dname}\label{sec:abl_dataset}
\paragraph{Influence of Multi-Layered Garments and Wind.}
As shown in \Tabref{tbl:splits},
we test models' abilities to animate garments on both simplified settings and general scenarios. The former assumes single-layered garments driven by human bodies, while the latter tests the models with multi-layered garments under the influences of both human bodies and wind.
Specifically,
we sample and divide our \dname~into four splits according to the types of garments and the strength of wind as indicated in \Tabref{tbl:splits}.
We train and test models with either only inner garments, which is marked by *,
or multi-layered garments on these four splits.
Notice that we group winds with a strength less than 50 as not windy,
where the wind has little influence on the garments.
Each split contains 36K frames for training,
2K frames for validation,
and 2K frames for test.
The training set, validation set, and test set are mutually exclusive,
thus they differ in human motions, garment topologies and attributes.

As shown in \Tabref{tbl:splits},
when DeePSD is trained with only inner garments,
 it achieves reasonable performance
compared with that when it is trained on Cloth3D \cite{DBLP:conf/eccv/BerticheME20}.
MGNet fails in our dataset due to the garment-specific design and low generalization abilities.
Our \mname~has lower errors, especially on splits with loose inner garments (L and L+W),
suggesting the effectiveness and higher generalization abilities
to animate loose clothes.
On the splits with wind (T+W and L+W),
DeePSD shows higher errors due to the random wind,
suggesting its poor generalization beyond human bodies.
Since jumpsuits in splits T and T+W are tight garments, the wind has less influence on them.

When trained on multi-layered garments, DeePSD's Euclidean errors increase, indicating that it struggles with modeling complex, layered clothing. In contrast, \mname~consistently demonstrates superior performance on all splits, handling both inner and outer garments effectively. The Euclidean errors remain similar across different splits, suggesting that our model exhibits greater robustness to varying garment topologies and external factors beyond human bodies.

\begin{figure}[h]
    \vspace{-2mm}
        \begin{center}
            \includegraphics[width=0.45\textwidth]{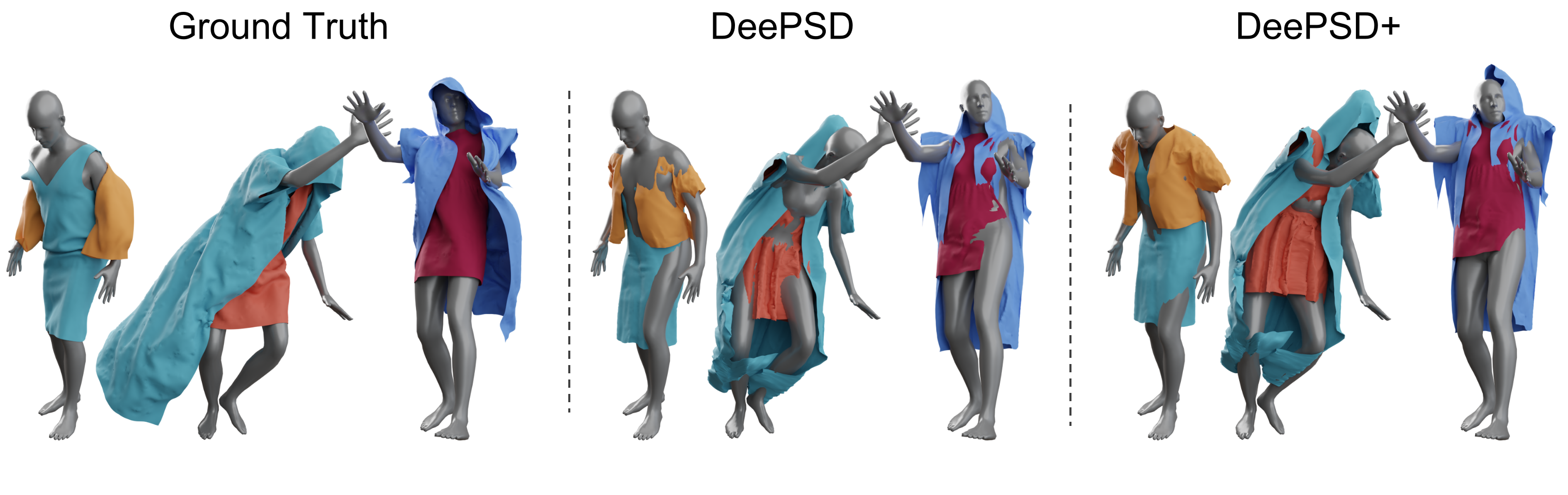}
        \end{center}
        \vspace{-6mm}
        \caption{\small{
        Samples of qualitative results by DeePSD.
        DeePSD has difficulties in capturing the rich dynamics in our \dname,
        leading to difficulties in convergence.
        With the collision loss,
        DeePSD+, which is mainly finetuned by collision loss on DeePSD,
        reduces part of interpenetration.
        However,
        DeePSD+ is not able to effectively animate multi-layred garments with diverse dynamics and complex topologies.
        The low rates of garment-to-garment interpenetration result from the interpenetration-free initialized garment templates. 
        }
        }
        \vspace{-6mm}
        \label{fig:baseline}
\end{figure}
\noindent{\textbf{General Garment Animations.}}
As demonstrated in \Tabref{tbl:full} and \Figref{fig:qualitative}, we further animate garments under more general conditions, featuring various combinations of multi-layered garments driven by human bodies and wind. Since DeePSD outperforms MGNet, we primarily compare our \mname~with DeePSD. For training, we uniformly sample 50K frames from \dname, along with 6K frames for validation and 6K frames for testing. There is no overlap among these sample sets. All samples include both inner and outer garments, as well as random wind as an external factor.

As shown in \Tabref{tbl:full}, the basic DeePSD without collision loss exhibits high Euclidean errors across all garment types. The intricate dynamics introduced by multi-layered garments and wind disrupt DeePSD, causing convergence difficulties as depicted in \Figref{fig:baseline}. As a result, DeePSD fails to accurately predict the garments' lively movements and leads to extensive garment-to-body collisions and garment-to-garment interpenetration. Although DeePSD+, which is fine-tuned with collision loss, attempts to resolve some of the collisions, it performs worse in terms of Euclidean errors. The relatively low collision rates between garment layers stem from the interpenetration-free initialization of garment templates. This feature allows DeePSD to automatically avoid some collisions when using linear blend skinning to deform the templates.

In contrast, \mname~delivers superior performance in terms of Euclidean errors and collision rates, demonstrating the effectiveness of our simulation-based formulation powered by rotation equivalent transformation. Our method also shows outstanding generalization for various garment types. Notably, \mname~achieves low collision rates and small Euclidean errors without penalizing collisions explicitly, resulting in more accurate outcomes. Since the core concept of simulation involves modeling object interactions, such as energy transitions and collisions, \mname~can resolve collisions implicitly. By incorporating collision loss, \mname~further minimizes interpenetration and strikes a balance between Euclidean errors and collision rates. We display the qualitative results of our \mname~in \Figref{fig:qualitative}. Additional qualitative comparisons can be found in the supplementary materials.

\begin{table}[t]
    \vspace{-2mm}
    \caption{
    \small{
    Ablation studies in terms of euclidean error (mm) and collision rates.
    We analyze the effectiveness of our Rotation Equivalent Transformation (RET),
    and the impacts of different collision loss.
    \mname~with default loss term $\mathcal{L}_{c,b}$,$\mathcal{L}_{c,g}$ adopts weights $\lambda_b=1.0, \lambda_g=0.1$,
    while \mname~with optimal combinations of $\mathcal{L}^{*}_{c,b}$,$\mathcal{L}^{*}_{c,g}$ adopts weight $\lambda_b=1.3$. 
    RET enables \mname~to reduce collisions especially garment-to-body collisions,
    making a balance between faithful predictions and low error rates.
    }}
    \vspace{-4mm}
    \label{tbl:model_abl}
    \setlength{\tabcolsep}{1.8pt}
    \begin{center}
    \small
    \begin{tabular}{lccc}
    \toprule
    \bf \mname	    & \bf Overall 	    & \bf L-Collision(\%)	& \bf H-Collision(\%)
    \\ \midrule
    Vanilla	&472.8$\pm$330.7 	& 3.13$\pm$2.22    &10.68$\pm$4.53\\
    + $\mathcal{L}_{c,b}$	&\textbf{446.3$\pm$304.9} 	& 3.10$\pm$2.16    &9.51$\pm$4.55\\
    + RET, $\mathcal{L}_{c,b}$          &449.2$\pm$315.1	& 5.21$\pm$3.34        &2.72$\pm$1.60\\
    + RET, $\mathcal{L}_{c,b}$,$\mathcal{L}_{c,g}$          &466.3$\pm$333.3	& \textbf{2.62$\pm$2.28}     &4.28$\pm$2.09\\
    + RET, $\mathcal{L}^{*}_{c,b}$,$\mathcal{L}^{*}_{c,g}$          &467.2$\pm$330.7	& 3.77$\pm$2.60     &\textbf{2.16$\pm$1.46}\\\bottomrule
    \end{tabular}
    \end{center}
    \vspace{-7mm}
\end{table}

\noindent{\textbf{Ablation Study.}}
We investigate the effectiveness of our Rotation Equivalent Transformation (RET) and the impact of various collision losses in \Tabref{tbl:model_abl}. \mname~with garment-to-human collision loss attains lower Euclidean errors and reduces collision rates between garments and human bodies. When trained with RET, \mname~dramatically decreases garment-to-human penetration by over 71\%, from 9.51\% to 2.72\%, while maintaining low Euclidean errors and garment-to-garment collision rates. This suggests that RET effectively eliminates redundant information from different rotations and enhances \mname's ability to capture the semantics of complex interactions. Due to its modeling of particle-wise interactions, which implicitly accounts for collisions, \mname~still achieves relatively low garment-to-garment penetration rates without $\mathcal{L}{c, g}$. Although $\mathcal{L}{c, g}$ slightly increases garment-to-body penetrations, an optimal combination of $\mathcal{L}{c, g}$ and $\mathcal{L}{c, b}$ jointly benefits \mname~, producing accurate predictions with low errors.

\section{Conclusion} \label{conclusion}

In this paper, we introduce a Transformer-based simulation method, named \mname, designed to animate diverse garments represented by patch-wise particles within a two-level hierarchical structure. The newly proposed Rotation Equivalent Transformation leverages the rotation equivariance and additivity of physics systems, enabling \mname~to effectively generalize across various garment animation scenarios. Moreover, we propose a large-scale, novel garment animation dataset called \dname, aiming to bridge the gap between experimental environments and realistic situations. \dname~is a dynamic animation dataset governed by physical laws, encompassing 4,900 distinct combinations of multi-layered garments and a total of 700K frames, with sequences extending up to 600 frames in length. As demonstrated by our experiments, \mname~delivers superior, robust performance, showcasing compelling generalization capabilities.

{\small
\bibliographystyle{ieee_fullname}
\bibliography{egbib}
}

\end{document}